# Issues in Stacked Generalization

**Kai Ming Ting**                 KMTING@DEAKIN.EDU.AU
*School of Computing and Mathematics*
*Deakin University, Australia.*

**Ian H. Witten**                  IHW@CS.WAIKATO.AC.NZ
*Department of Computer Science*
*University of Waikato, New Zealand.*

## Abstract

Stacked generalization is a general method of using a high-level model to combine lower-level models to achieve greater predictive accuracy. In this paper we address two crucial issues which have been considered to be a 'black art' in classification tasks ever since the introduction of stacked generalization in 1992 by Wolpert: the type of generalizer that is suitable to derive the higher-level model, and the kind of attributes that should be used as its input. We find that best results are obtained when the higher-level model combines the confidence (and not just the predictions) of the lower-level ones.

We demonstrate the effectiveness of stacked generalization for combining three different types of learning algorithms for classification tasks. We also compare the performance of stacked generalization with majority vote and published results of arcing and bagging.

## 1. Introduction

Stacked generalization is a way of combining multiple models that have been learned for a classification task (Wolpert, 1992), which has also been used for regression (Breiman, 1996a) and even unsupervised learning (Smyth & Wolpert, 1997). Typically, different learning algorithms learn different models for the task at hand, and in the most common form of stacking the first step is to collect the output of each model into a new set of data. For each instance in the original training set, this data set represents every model's prediction of that instance's class, along with its true classification. During this step, care is taken to ensure that the models are formed from a batch of training data that does not include the instance in question, in just the same way as ordinary cross-validation. The new data are treated as the data for another learning problem, and in the second step a learning algorithm is employed to solve this problem. In Wolpert's terminology, the original data and the models constructed for them in the first step are referred to as *level-0 data* and *level-0 models*, respectively, while the set of cross-validated data and the second-stage learning algorithm are referred to as *level-1 data* and the *level-1 generalizer*.

In this paper, we show how to make stacked generalization work for classification tasks by addressing two crucial issues which Wolpert (1992) originally described as 'black art' and have not been resolved since. The two issues are (i) the type of attributes that should be used to form level-1 data, and (ii) the type of level-1 generalizer in order to get improved accuracy using the stacked generalization method.

Breiman (1996a) demonstrated the success of stacked generalization in the setting of ordinary regression. The level-0 models are regression trees of different sizes or linear regressions using different number of variables. But instead of selecting the single model that works best as judged by (for example) cross-validation, Breiman used the different level-0 regressors' output values for each member of the training set to form level-1 data. Then he used least-squares linear regression, under the constraint that all regression coefficients be non-negative, as the level-1 generalizer. The non-negativity constraint turned out to be crucial to guarantee that the predictive accuracy would be better than that achieved by selecting the single best predictor.

Here we show how stacked generalization can be made to work reliably in classification tasks. We do this by using the output class probabilities generated by level-0 models to form level-1 data. Then for the level-1 generalizer we use a version of least squares linear regression adapted for classification tasks. We find the use of class probabilities to be crucial for the successful application of stacked generalization in classification tasks. However, the non-negativity constraints found necessary by Breiman in regression are found to be irrelevant to improved predictive accuracy in our classification situation.

In Section 2, we formally introduce the technique of stacked generalization and describe pertinent details of each learning algorithm used in our experiments. Section 3 describes the results of stacking three different types of learning algorithms. Section 4 compares stacked generalization with arcing and bagging, two recent methods that employ sampling techniques to modify the data distribution in order to produce multiple models from a single learning algorithm. The following section describes related work, and the paper ends with a summary of our conclusions.

## 2. Stacked Generalization

Given a data set $\mathcal{L} = \{(y_n, x_n), n = 1, \ldots, N\}$, where $y_n$ is the class value and $x_n$ is a vector representing the attribute values of the $n$th instance, randomly split the data into $J$ almost equal parts $\mathcal{L}_1, \ldots, \mathcal{L}_J$. Define $\mathcal{L}_j$ and $\mathcal{L}^{(-j)} = \mathcal{L} - \mathcal{L}_j$ to be the test and training sets for the $j$th fold of a $J$-fold cross-validation. Given $K$ learning algorithms, which we call *level-0 generalizers*, invoke the $k$th algorithm on the data in the training set $\mathcal{L}^{(-j)}$ to induce a model $\mathcal{M}_k^{(-j)}$, for $k = 1, \ldots, K$. These are called *level-0 models*.

For each instance $x_n$ in $\mathcal{L}_j$, the test set for the $j$th cross-validation fold, let $z_{kn}$ denote the prediction of the model $\mathcal{M}_k^{(-j)}$ on $x_n$. At the end of the entire cross-validation process, the data set assembled from the outputs of the $K$ models is

$$\mathcal{L}_{CV} = \{(y_n, z_{1n}, \ldots, z_{Kn}), n = 1, \ldots, N\}.$$

These are the *level-1 data*. Use some learning algorithm that we call the *level-1 generalizer* to derive from these data a model $\tilde{\mathcal{M}}$ for $y$ as a function of $(z_1, \ldots, z_K)$. This is the *level-1 model*. Figure 1 illustrates the cross-validation process. To complete the training process, the final level-0 models $\mathcal{M}_k$, $k = 1, \ldots, K$, are derived using all the data in $\mathcal{L}$.

Now let us consider the classification process, which uses the models $\mathcal{M}_k$, $k = 1, \ldots, K$, in conjunction with $\tilde{\mathcal{M}}$. Given a new instance, models $\mathcal{M}_k$ produce a vector $(z_1, \ldots, z_K)$. This vector is input to the level-1 model $\tilde{\mathcal{M}}$, whose output is the final classification result for that instance. This completes the stacked generalization method as proposed by Wolpert (1992), and also used by Breiman (1996a) and LeBlanc & Tibshirani (1993).

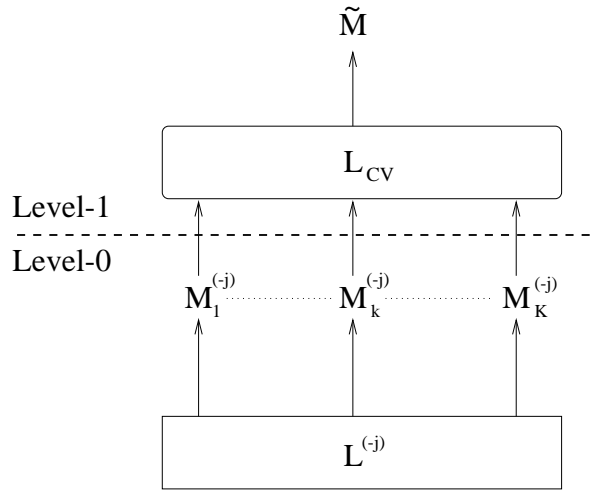

Figure 1: This figure illustrates the $j$-fold cross-validation process in level-0; and the level-1 data set $\mathcal{L}_{CV}$ at the end of this process is used to produce level-1 model $\tilde{\mathcal{M}}$.

As well as the situation described above, which results in the level-1 model $\tilde{\mathcal{M}}$, the present paper also considers a further situation where the output from the level-0 models is a set of class probabilities rather than a single class prediction. If model $\mathcal{M}_k^{(-j)}$ is used to classify an instance $x$ in $\mathcal{L}_j$, let $P_{ki}(x)$ denote the probability of the $i$th output class, and the vector

$$\mathcal{P}_{kn} = (P_{k1}(x_n), \ldots, P_{ki}(x_n), \ldots, P_{kI}(x_n))$$

gives the model's class probabilities for the $n$th instance, assuming that there are $I$ classes. As the level-1 data, assemble together the class probability vector from all $K$ models, along with the actual class:

$$\mathcal{L}'_{CV} = \{(y_n, \mathcal{P}_{1n}, \ldots, \mathcal{P}_{kn}, \ldots, \mathcal{P}_{Kn}), n = 1, \ldots, N\}.$$

Denote the level-1 model derived from this as $\tilde{\mathcal{M}}'$ to contrast it with $\tilde{\mathcal{M}}$.

The following two subsections describe the algorithms used as level-0 and level-1 generalizers in the experiments reported in Section 3.

## 2.1 Level-0 Generalizers

Three learning algorithms are used as the level-0 generalizers: C4.5, a decision tree learning algorithm (Quinlan, 1993); NB, a re-implementation of a Naive Bayesian classifier (Cestnik, 1990); and IB1, a variant of a lazy learning algorithm (Aha, Kibler & Albert, 1991) which employs the $p$-nearest-neighbor method using a modified value-difference metric for nominal and binary attributes (Cost & Salzberg, 1993). For each of these learning algorithms we now show the formula that we use for the estimated output class probabilities $P_i(x)$ for an instance $x$ (where, in all cases, $\sum_i P_i(x) = 1$).

**C4.5:** Consider the leaf of the decision tree at which the instance $x$ falls. Let $m_i$ be the number of (training) instances with class $i$ at this leaf, and suppose the majority class at the leaf is $\hat{I}$. Let $E = \sum_{i \neq \hat{i}} m_i$. Then, using a Laplace estimator,

$$P_{\hat{I}}(x) = 1 - \frac{E + 1}{\sum_i m_i + 2},$$

$$P_i(x) = (1 - P_{\hat{I}}(x)) \times \frac{m_i}{E}, \text{ for } i \neq \hat{I}.$$

Note that only pruned trees and default settings of C4.5 are used in our experiments.

**NB:** Let $P(i|x)$ be the posterior probability of class $i$, given instance $x$. Then

$$P_i(x) = \frac{P(i|x)}{\sum_i P(i|x)}.$$

Note that NB uses a Laplacian estimate for estimating the conditional probabilities for each nominal attribute to compute $P(i|x)$. For each continuous-valued attribute, a normal distribution is assumed in which case the conditional probabilities can be conveniently represented entirely in terms of the mean and variance of the observed values for each class.

**IB1:** Suppose $p$ nearest neighbors are used; denote them by $\{(y_s, x_s), s = 1, \ldots, p\}$ for instance $x$. (We use $p = 3$ in the experiments.) Then

$$P_i(x) = \frac{\sum_{s=1}^{p} f(y_s)/d(x, x_s)}{\sum_{s=1}^{p} 1/d(x, x_s)},$$

where $f(y_s) = 1$ if $i = y_s$ and 0 otherwise, and $d$ is the Euclidean distance function.

In all three learning algorithms, the predicted class of the level-0 model, given an instance $x$, is that $\hat{I}$ for which

$$P_{\hat{I}}(x) > P_i(x) \text{ for all } i \neq \hat{I}.$$

## 2.2 Level-1 Generalizers

We compare the effect of four different learning algorithms as the level-1 generalizer: C4.5, IB1(using $p = 21$ nearest neighbors),[1] NB, and a multi-response linear regression algorithm, MLR. Only the last needs further explanation.

MLR is an adaptation of a least-squares linear regression algorithm that Breiman (1996a) used in regression settings. Any classification problem with real-valued attributes can be transformed into a multi-response regression problem. If the original classification problem has $I$ classes, it is converted into $I$ separate regression problems, where the problem for class $\ell$ has instances with responses equal to one when they have class $\ell$ and zero otherwise.

The input to MLR is level-1 data, and we need to consider the situation for the model $\tilde{\mathcal{M}}'$, where the attributes are probabilities, separately from that for the model $\tilde{\mathcal{M}}$, where they are classes. In the former case, where the attributes are already real-valued, the linear regression for class $\ell$ is simply

$$LR_\ell(x) = \sum_k^K \alpha_{k\ell} P_{k\ell}(x).$$

---

1. A large $p$ value is used following Wolpert's (1992) advice that "...it is reasonable that 'relatively global, smooth ...' level-1 generalizers should perform well."

In the latter case, the classes are unordered nominal attributes. We map them into binary values in the obvious way, setting $P_{k\ell}(x)$ to 1 if the class of instance $x$ is $\ell$ and zero otherwise; and then use the above linear regression.

Choose the linear regression coefficients $\{\alpha_{k\ell}\}$ to minimize

$$\sum_j \sum_{(y_n, x_n) \in \mathcal{L}_j} (y_n - \sum_k \alpha_{k\ell} P_{k\ell}^{(-j)}(x_n))^2.$$

The coefficients $\{\alpha_{k\ell}\}$ are constrained to be non-negative, following Breiman's (1996a) discovery that this is necessary for the successful application of stacked generalization to regression problems. The non-negative-coefficient least-squares algorithm described by Lawson & Hanson (1995) is employed here to derive the linear regression for each class. We show later that, in fact, the non-negative constraint is unnecessary in classification tasks.

With this in place, we can now describe the working of MLR. To classify a new instance $x$, compute $LR_\ell(x)$ for all $I$ classes and assign the instance to that class $\ell$ which has the greatest value:[2]

$$LR_\ell(x) > LR_{\ell'}(x) \text{ for all } \ell' \neq \ell.$$

In the next section we investigate the stacking of C4.5, NB and IB1.

## 3. Stacking C4.5, NB and IB1

### 3.1 When Does Stacked Generalization Work?

The experiments in this section show that

- for successful stacked generalization it is necessary to use output class probabilities rather than class predictions—that is, $\tilde{\mathcal{M}}'$ rather than $\tilde{\mathcal{M}}$;

- only the MLR algorithm is suitable for the level-1 generalizer, among the four algorithms used.

We use two artificial datasets and eight real-world datasets from the UCI Repository of machine learning databases (Blake, Keogh & Merz, 1998). Details of these are given in Table 1.

For the artificial datasets—Led24 and Waveform—each training dataset $\mathcal{L}$ of size 200 and 300, respectively, is generated using a different seed. The algorithms used for the experiments are then tested on a separate dataset of 5000 instances. Results are expressed as the average error rate of ten repetitions of this entire procedure.

For the real-world datasets, $W$-fold cross-validation is performed. In each fold of this cross-validation, the training dataset is used as $\mathcal{L}$, and the models derived are evaluated

---

2. The pattern recognition community calls this type of classifier a *linear machine* (Duda & Hart, 1973).

| Datasets | # Samples | # Classes | # Attr & Type |
|----------|-----------|-----------|---------------|
| Led24 | 200/5000 | 10 | 10N |
| Waveform | 300/5000 | 3 | 40C |
| Horse | 368 | 2 | 3B+12N+7C |
| Credit | 690 | 2 | 4B+5N+6C |
| Vowel | 990 | 11 | 10C |
| Euthyroid | 3163 | 2 | 18B+7C |
| Splice | 3177 | 3 | 60N |
| Abalone | 4177 | 3 | 1N+7C |
| Nettalk(s) | 5438 | 5 | 7N |
| Coding | 20000 | 2 | 15N |

N-nominal; B-binary; C-Continuous.

Table 1: Details of the datasets used in the experiment.

on the test dataset. The result is expressed as the average error rate of the $W$-fold cross-validation. Note that this cross-validation is used for evaluation of the entire procedure, whereas the $J$-fold cross-validation mentioned in Section 2 is the internal operation of stacked generalization. However, both $W$ and $J$ are set to 10 in the experiments.

In this section, we present results of model combination using level-1 models $\tilde{\mathcal{M}}$ and $\tilde{\mathcal{M}}'$, as well as a model selection method, employing the same $J$-fold cross-validation procedure. Note that the only difference between model combination and model selection here is whether the level-1 learning is employed or not.

Table 2 shows the average error rates, obtained using $W$-fold cross-validation, of C4.5, NB and IB1, and BestCV, which is the best of the three, selected using $J$-fold cross-validation. As expected, BestCV is almost always the classifier with the lowest error rate.[3]

Table 3 shows the result of stacked generalization using the level-1 model $\tilde{\mathcal{M}}$, for which the level-1 data comprise the classifications generated by the level-0 models, and $\tilde{\mathcal{M}}'$, for which the level-1 data comprise the probabilities generated by the level-0 models. Results are shown for all four level-1 generalizers in each case, along with BestCV. The lowest error rate for each dataset is given in bold.

Table 4 summarizes the results in Table 3 in terms of a comparison of each level-1 model with BestCV totaled over all datasets. Clearly, the best level-1 model is $\tilde{\mathcal{M}}'$ derived using MLR. It performs better than BestCV in nine datasets and equally well in the tenth. The best performing $\tilde{\mathcal{M}}$ is derived from NB, which performs better than BestCV in seven datasets but significantly worse in two (Waveform and Vowel). We regard a difference of more than two standard errors as significant (95% confidence). The standard error figures are omitted in this table to increase readability.

The datasets are shown in the order of increasing size. MLR performs significantly better than BestCV in the four largest datasets. This indicates that stacked generalization is more likely to give significant improvements in predictive accuracy if the volume of data is large—a direct consequence of more accurate estimation using cross-validation.

---

3. Note that BestCV does not always select the same classifier in all $W$ folds. That is why its error rate is not always equal to the lowest error rate among the three classifiers.

| Datasets | Level-0 Generalizers | | | |
|----------|------|------|------|--------|
| | C4.5 | NB | IB1 | BestCV |
| Led24 | 35.4 | 35.4 | 32.2 | 32.8 ±0.6 |
| Waveform | 31.8 | 17.1 | 26.2 | 17.1 ±0.3 |
| Horse | 15.8 | 17.9 | 15.8 | 17.1 ±1.6 |
| Credit | 17.4 | 17.3 | 28.1 | 17.4 ±1.2 |
| Vowel | 22.7 | 51.0 | 2.6 | 2.6 ±0.2 |
| Euthyroid | 1.9 | 9.8 | 8.6 | 1.9 ±0.3 |
| Splice | 5.5 | 4.5 | 4.7 | 4.5 ±0.4 |
| Abalone | 41.4 | 42.1 | 40.5 | 40.1 ±0.6 |
| Nettalk(s) | 17.0 | 15.9 | 12.7 | 12.7 ±0.4 |
| Coding | 27.6 | 28.8 | 25.0 | 25.0 ±0.3 |

Table 2: Average error rates of C4.5, NB and IB1, and BestCV—the best among them selected using $J$-fold cross-validation. The standard errors are shown in the last column.

| Datasets | | Level-1 model, $\mathcal{M}$ | | | | Level-1 model, $\mathcal{M}'$ | | | |
|----------|--------|------|------|------|------|------|------|------|------|
| | BestCV | C4.5 | NB | IB1 | MLR | C4.5 | NB | IB1 | MLR |
| Led24 | 32.8 | 34.0 | 32.4 | 35.0 | 33.3 | 41.7 | 35.7 | 32.1 | **31.3** |
| Waveform | 17.1 | 17.7 | 19.2 | 18.7 | 17.2 | 20.6 | 17.6 | 17.8 | **16.8** |
| Horse | 17.1 | 16.9 | **14.9** | 17.6 | 16.3 | 18.0 | 18.5 | 17.7 | 15.2 |
| Credit | 17.4 | 18.4 | 16.1 | 16.9 | 17.4 | 15.4 | 15.9 | **14.3** | 16.2 |
| Vowel | 2.6 | 2.6 | 3.8 | 3.6 | 2.6 | 2.7 | 7.2 | 3.3 | **2.5** |
| Euthyroid | **1.9** | **1.9** | **1.9** | **1.9** | **1.9** | 2.2 | 4.3 | 2.0 | **1.9** |
| Splice | 4.5 | 3.9 | 3.9 | **3.8** | **3.8** | 4.0 | 3.9 | **3.8** | **3.8** |
| Abalone | 40.1 | 38.5 | 38.5 | 38.2 | 38.1 | 43.3 | **37.1** | 39.2 | 38.3 |
| Nettalk(s) | 12.7 | 12.4 | 11.9 | 12.4 | 12.6 | 14.0 | 14.6 | 12.0 | **11.5** |
| Coding | 25.0 | 23.2 | 23.1 | 23.2 | 23.2 | 22.3 | 21.2 | 21.2 | **20.7** |

Table 3: Average error rates for stacking C4.5, NB and IB1.

| | Level-1 model, $\mathcal{M}$ | | | | Level-1 model, $\mathcal{M}'$ | | | |
|----------|------|------|------|------|------|------|------|------|
| | C4.5 | NB | IB1 | MLR | C4.5 | NB | IB1 | MLR |
| #Win vs. #Loss | 3-5 | 2-7 | 4-5 | 2-5 | 7-3 | 6-4 | 4-6 | 0-9 |

Table 4: Summary of Table 3—Comparison of BestCV with $\tilde{\mathcal{M}}$ and $\tilde{\mathcal{M}}'$.

When one of the level-0 models performs significantly much better than the rest, like in the Euthyroid and Vowel datasets, MLR performs either as good as BestCV by selecting the best performing level-0 model, or better than BestCV.

MLR has an advantage over the other three level-1 generalizers in that its model can easily be interpreted. Examples of the combination weights it derives (for the probability-based model $\mathcal{M}'$) appear in Table 5 for the Horse, Credit, Splice, Abalone, Waveform, Led24 and Vowel datasets. The weights indicate the relative importance of the level-0 generalizers for each prediction class. For example, in the Splice dataset (in Table 5(b)), NB is the dominant generalizer for predicting class 2, NB and IB1 are both good at predicting class 3, and all three generalizers make a worthwhile contribution to the prediction of class 1. In contrast, in the Abalone dataset all three generalizers contribute substantially to the prediction of all three classes. Note that the weights for each class do not sum to one because no such constraint is imposed on MLR.

## 3.2 Are Non-negativity Constraints Necessary?

Both Breiman (1996a) and LeBlanc & Tibshirani (1993) use the stacked generalization method in a regression setting and report that it is necessary to constrain the regression coefficients to be non-negative in order to guarantee that stacked regression improves predictive accuracy. Here we investigate this finding in the domain of classification tasks.

To assess the effect of the non-negativity constraint on performance, three versions of MLR are employed to derive the level-1 model $\mathcal{M}'$:

i. each linear regression in MLR is calculated with an intercept constant (that is, $I+1$ weights for the $I$ classes) but without any constraints;

ii. each linear regression is derived with neither an intercept constant ($I$ weights for $I$ classes) nor constraints;

iii. each linear regression is derived without an intercept constant, but with non-negativity constraints ($I$ non-negative weights for $I$ classes).

The third version is the one used for the results presented earlier. Table 6 shows the results of all three versions. *They all have almost indistinguishable error rates.* We conclude that in classification tasks, non-negativity constraints are not necessary to guarantee that stacked generalization improves predictive accuracy.

However, there is another reason why it is a good idea to employ non-negativity constraints. Table 7 shows an example of the weights derived by these three versions of MLR on the Led24 dataset. The third version, shown in column (iii), supports a more perspicuous interpretation of each level-0 generalizer's contribution to the class predictions than do the other two. In this dataset, IB1 is the dominant generalizer in predicting classes 4, 5 and 8, and both NB and IB1 make a worthwhile contribution in predicting class 2, as evidenced by their high weights. However, the negative weights used in predicting these classes render the interpretation of the other two versions much less clear.

| Class | Horse | | | Credit | | |
|---|---|---|---|---|---|---|
| | C4.5 | NB | IB1 | C4.5 | NB | IB1 |
| 1 | 0.36 | 0.20 | 0.42 | 0.63 | 0.30 | 0.04 |
| 2 | 0.39 | 0.19 | 0.41 | 0.65 | 0.28 | 0.07 |

C4.5 for $\alpha_1$; NB for $\alpha_2$; IB1 for $\alpha_3$.

Table 5: (a) Weights generated by MLR (model $\bar{\mathcal{M}}'$) for the Horse and Credit datasets.

| Class | Splice | | | Abalone | | | Waveform | | |
|---|---|---|---|---|---|---|---|---|---|
| | C4.5 | NB | IB1 | C4.5 | NB | IB1 | C4.5 | NB | IB1 |
| 1 | 0.23 | 0.43 | 0.36 | 0.25 | 0.25 | 0.39 | 0.16 | 0.59 | 0.34 |
| 2 | 0.15 | 0.72 | 0.12 | 0.27 | 0.20 | 0.25 | 0.14 | 0.72 | 0.07 |
| 3 | 0.08 | 0.52 | 0.40 | 0.30 | 0.18 | 0.39 | 0.04 | 0.65 | 0.23 |

Table 5: (b) Weights generated by MLR (model $\bar{\mathcal{M}}'$) for the Splice, Abalone and Waveform datasets.

| Class | Led24 | | | Vowel | | |
|---|---|---|---|---|---|---|
| | C4.5 | NB | IB1 | C4.5 | NB | IB1 |
| 1 | 0.46 | 0.65 | 0.00 | 0.04 | 0.00 | 0.96 |
| 2 | 0.00 | 0.37 | 0.43 | 0.03 | 0.00 | 0.97 |
| 3 | 0.47 | 0.00 | 0.54 | 0.01 | 0.00 | 1.00 |
| 4 | 0.00 | 0.13 | 0.65 | 0.05 | 0.25 | 0.86 |
| 5 | 0.00 | 0.19 | 0.64 | 0.01 | 0.08 | 0.97 |
| 6 | 0.35 | 0.14 | 0.35 | 0.15 | 0.00 | 0.92 |
| 7 | 0.15 | 0.43 | 0.36 | 0.03 | 0.01 | 1.02 |
| 8 | 0.00 | 0.00 | 0.68 | 0.04 | 0.01 | 0.96 |
| 9 | 0.00 | 0.38 | 0.29 | 0.03 | 0.00 | 1.02 |
| 10 | 0.00 | 0.50 | 0.24 | 0.08 | 0.01 | 0.93 |
| 11 | – | – | – | 0.00 | 0.04 | 0.96 |

Table 5: (c) Weights generated by MLR (model $\bar{\mathcal{M}}'$) for the Led24 and Vowel datasets.

| Datasets | MLR with | | |
|----------|----------|---|---|
| | No Constraints | No Intercept | Non-Negativity |
| Led24 | 31.4 | 31.4 | 31.3 |
| Waveform | 16.8 | 16.8 | 16.8 |
| Horse | 15.8 | 15.8 | 15.2 |
| Credit | 16.2 | 16.2 | 16.2 |
| Vowel | 2.4 | 2.4 | 2.5 |
| Euthyroid | 1.9 | 1.9 | 1.9 |
| Splice | 3.7 | 3.8 | 3.8 |
| Abalone | 38.3 | 38.3 | 38.3 |
| Nettalk(s) | 11.6 | 11.5 | 11.5 |
| Coding | 20.7 | 20.7 | 20.7 |

Table 6: Average error rates of three versions of MLR.

| | (i) | | | | (ii) | | | (iii) | | |
|---|---|---|---|---|---|---|---|---|---|---|
| Class | $\alpha_0$ | $\alpha_1$ | $\alpha_2$ | $\alpha_3$ | $\alpha_1$ | $\alpha_2$ | $\alpha_3$ | $\alpha_1$ | $\alpha_2$ | $\alpha_3$ |
| 1 | 0.00 | 0.45 | 0.65 | 0.00 | 0.46 | 0.65 | 0.00 | 0.46 | 0.65 | 0.00 |
| 2 | 0.02 | −0.42 | 0.47 | 0.56 | −0.40 | 0.49 | 0.56 | 0.00 | 0.37 | 0.43 |
| 3 | 0.00 | 0.46 | −0.01 | 0.54 | 0.47 | −0.01 | 0.54 | 0.47 | 0.00 | 0.54 |
| 4 | 0.04 | −0.33 | 0.15 | 0.84 | −0.29 | 0.21 | 0.81 | 0.00 | 0.13 | 0.65 |
| 5 | 0.03 | −0.37 | 0.26 | 0.84 | −0.32 | 0.26 | 0.84 | 0.00 | 0.19 | 0.64 |
| 6 | 0.01 | 0.35 | 0.12 | 0.35 | 0.36 | 0.14 | 0.35 | 0.35 | 0.14 | 0.35 |
| 7 | 0.01 | 0.15 | 0.43 | 0.36 | 0.15 | 0.43 | 0.36 | 0.15 | 0.43 | 0.36 |
| 8 | 0.02 | −0.05 | −0.25 | 0.72 | −0.03 | −0.19 | 0.72 | 0.00 | 0.00 | 0.68 |
| 9 | 0.04 | −0.08 | 0.32 | 0.32 | −0.05 | 0.40 | 0.30 | 0.00 | 0.38 | 0.29 |
| 10 | 0.04 | −0.06 | 0.43 | 0.25 | −0.01 | 0.50 | 0.24 | 0.00 | 0.50 | 0.24 |

Table 7: Weights generated by three versions of MLR: (i) no constraints, (ii) no intercept, and (iii) non-negativity constraints, for the LED24 dataset.

| Dataset | #SE | BestCV | Majority | MLR |
|---------|-----|--------|----------|-----|
| Horse | 0.5 | 17.1 | **15.0** | 15.2 |
| Splice | 2.5 | 4.5 | 4.0 | **3.8** |
| Abalone | 3.3 | 40.1 | 39.0 | **38.3** |
| Led24 | 8.7 | 32.8 | 31.8 | **31.3** |
| Credit | 8.9 | 17.4 | **16.1** | 16.2 |
| Nettalk(s) | 10.8 | 12.7 | 12.2 | **11.5** |
| Coding | 12.7 | 25.0 | 23.1 | **20.7** |
| Waveform | 18.7 | 17.1 | 19.5 | **16.8** |
| Euthyroid | 26.3 | **1.9** | 8.1 | **1.9** |
| Vowel | 242.0 | 2.6 | 13.0 | **2.5** |

Table 8: Average error rates of BestCV, Majority Vote and MLR (model $\tilde{\mathcal{M}}'$), along with the number of standard error (#SE) between BestCV and the worst performing level-0 generalizers.

## 3.3 How Does Stacked Generalization Compare To Majority Vote?

Let us now compare the error rate of $\tilde{\mathcal{M}}'$, derived from MLR, to that of majority vote, a simple decision combination method which requires neither cross-validation nor level-1 learning. Table 8 shows the average error rates of BestCV, majority vote and MLR. In order to see whether the relative performances of level-0 generalizers have any effect on these methods, the number of standard errors (#SE) between the error rates of the worst performing level-0 generalizer and BestCV is given, and the datasets are re-ordered according to this measure. Since BestCV almost always selects the best performing level-0 generalizer, small values of #SE indicate that the level-0 generalizers perform comparably to one another, and vice versa.

MLR compares favorably to majority vote, with eight wins versus two losses. Out of the eight wins, six have significant differences (the two exceptions are for the Splice and Led24 datasets); whereas both losses (for the Horse and Credit datasets) have insignificant differences. Thus the extra computation for cross-validation and level-1 learning seems to have paid off.

It is interesting to note that the performance of majority vote is related to the size of #SE. Majority vote compares favorably to BestCV in the first seven datasets, where the values of #SE are small. In the last three, where #SE is large, majority vote performs worse. This indicates that if the level-0 generalizers perform comparably, it is not worth using cross-validation to determine the best one, because the result of majority vote—which is far cheaper—is not significantly different. Although small values of #SE are a necessary condition for majority vote to rival BestCV, they are not a sufficient condition—see Matan (1996) for an example. The same applies when majority vote is compared with MLR. MLR performs significantly better in the five datasets that have large #SE values, but in only one of the other cases.

| | $\mathcal{M}$ versus $\mathcal{M}'$ | | | |
|---|---|---|---|---|
| | C4.5 | NB | IB1 | MLR |
| #Win vs. #Loss | 8-2 | 5-4 | 3-6 | 1-7 |

Table 9: $\tilde{\mathcal{M}}$ versus $\tilde{\mathcal{M}}'$ for each generalizer—summarized results from Table 3.

It is worth mentioning a method that averages $P_i(x)$ for each $i$ over all level-0 models, yielding $\bar{P}_i(x)$, and then predicts class $\hat{I}$ for which $\bar{P}_{\hat{I}}(x) > \bar{P}_i(x)$ for all $i \neq \hat{I}$. According to Breiman (1996b), this method produces an error rate almost identical to that of majority vote.

## 3.4 Why Does Stacked Generalization Work Best With $\tilde{\mathcal{M}}'$ Generated From MLR?

We have shown that stacked generalization works best when output class probabilities (rather than class predictions) are used with the MLR algorithm (rather than C4.5, IB1, NB). In retrospect, this is not surprising, and can be explained intuitively as follows. The level-1 model should provide a simple way of combining all the evidence available. This evidence includes not just the predictions, but the confidence of each level-0 model in its predictions. A linear combination is the simplest way of pooling the level-0 models' confidence, and MLR provides just that.

The alternative methods of NB, C4.5, and IB1 each have shortcomings. A Bayesian approach could form the basis for a suitable alternative way of pooling the level-0 models' confidence, but the independence assumption central to Naive Bayes hampers its performance in some datasets because the evidence provided by the individual level-0 models is certainly not independent. C4.5 builds trees that can model interaction amongst attributes—particularly when the tree is large—but this is not desirable for combining confidences. Nearest neighbor methods do not really give a way of combining confidences; also, the similarity metric employed could misleadingly assume that two different sets of confidence levels are similar.

Table 9 summarizes the results in Table 3 by comparing $\tilde{\mathcal{M}}$ with $\tilde{\mathcal{M}}'$ for each level-1 generalizer, across all datasets. C4.5 is clearly better off with a label-based representation, because discretizing continuous-valued attributes creates intra-attribute interaction in addition to interactions between different attributes. The evidence from Table 9 is that NB is indifferent to the use of labels or confidences: the normal distribution assumption that it embodies in the latter case could be another reason why it is unsuitable for combining confidence measures. Both MLR and IB1 handle continuous-valued attributes better than label-based ones, since this is the domain in which they are designed to work.

## Summary

We summarize our findings in this section as follows.

- None of the four learning algorithms used to obtain model $\tilde{\mathcal{M}}$ perform satisfactorily.

- MLR is the best of the four learning algorithms to use as the level-1 generalizer for obtaining the model $\tilde{\mathcal{M}}'$.

- When obtained using MLR, $\tilde{\mathcal{M}}'$ has lower predictive error rate than the best model selected by $J$-fold cross-validation, for almost all datasets used in the experiments.

- Another advantage of MLR over the other three level-1 generalizers is its interpretability. The weights $\alpha_{k\ell}$ indicate the different contributions that each level-0 model $k$ makes to the prediction classes $\ell$.

- Model $\tilde{\mathcal{M}}'$ can be derived by MLR with or without non-negativity constraints. Such constraints make little difference to the model's predictive accuracy.

- The use of non-negativity constraints in MLR has the advantage of interpretability. Non-negative weights $\alpha_{k\ell}$ support easier interpretation of the extent to which each model contributes to each prediction class.

- When derived using MLR, model $\tilde{\mathcal{M}}'$ compares favorably with majority vote.

- MLR provides a method of combining the confidence generated by the level-0 models into a final decision. For various reasons, NB, C4.5, and IB1 are not suitable for this task.

## 4. Comparison With Arcing And Bagging

This section compares the results of stacking C4.5, NB and IB1 with the results of arcing (called boosting by its originator, Schapire, 1990) and bagging that are reported by Breiman (1996b; 1996c). Both arcing and bagging employ sampling techniques to modify the data distribution in order to produce multiple models from a single learning algorithm. To combine the decisions of the individual models, arcing uses a weighted majority vote and bagging uses an unweighted majority vote. Breiman reports that both arcing and bagging can substantially improve the predictive accuracy of a single model derived using a base learning algorithm.

### 4.1 Experimental Results

First we describe the differences between the experimental procedures. Our results for stacking are averaged over ten-fold cross-validation for all datasets except Waveform, which is averaged over ten repeated trials. Standard errors are also shown. Results for arcing and bagging are those obtained by Breiman (1996b; 1996c), which are averaged over 100 trials. In Breiman's experiments, each trial uses a random 9:1 split to form the training and test sets for all datasets except Waveform. Also note that the Waveform dataset we used has 19 irrelevant attributes, but Breiman used a version without irrelevant attributes (which would be expected to degrade the performance of level-0 generalizers in our experiments). In both cases 300 training instances were used for this dataset, but we used 5000 test instances whereas Breiman used 1800. Arcing and bagging are done with 50 decision tree models derived from CART (Breiman *et al.*, 1984) in each trial.

| Dataset | #Samples | stacking | arcing | bagging |
|---------|----------|----------|--------|---------|
| Waveform | 300 | 16.8 ±0.2 | 17.8 | 19.3 |
| Glass | 214 | 28.4 ±2.9 | 22.0 | 23.2 |
| Ionosphere | 351 | 9.7 ±1.5 | 6.4 | 7.9 |
| Soybean | 683 | 4.3 ±1.1 | 5.8 | 6.8 |
| Breast Cancer | 699 | 2.7 ±0.8 | 3.2 | 3.7 |
| Diabetes | 768 | 24.2 ±1.2 | 26.6 | 23.9 |

Table 10: Comparing stacking with arcing and bagging classifiers.

The results on six datasets are given in Table 10, and indicate that the three methods are very competitive.[4] Stacking performs better than both arcing and bagging in three datasets (Waveform, Soybean and Breast Cancer), and is better than arcing but worse than bagging in the Diabetes dataset. Note that stacking performs very poorly on Glass and Ionosphere, two small real-world datasets. This is not surprising, because cross-validation inevitably produces poor estimates for small datasets.

## 4.2 Discussion

Like bagging, stacking is ideal for parallel computation. The construction of each level-0 model proceeds independently, no communication with the other modeling processes being necessary.

Arcing and bagging require a considerable number of member models because they rely on varying the data distribution to get a diverse set of models from a single learning algorithm. Using a level-1 generalizer, stacking can work with only two or three level-0 models.

Suppose the computation time required for a learning algorithm is $C$, and arcing or bagging needs $h$ models. The learning time required is $T_a = hC$. Suppose stacking requires $g$ models and each model employs $J$-fold cross-validation. Assuming that time $C$ is needed to derive each of the $g$ level-0 models and the level-1 model, the learning time for stacking is $T_s = (g(J + 1) + 1)C$. For the results given in Table 10, $h = 50$, $J = 10$, and $g = 3$; thus $T_a = 50C$ and $T_s = 34C$. However, in practice the learning time required for the level-0 and level-1 generalizers may be different.

Users of stacking have a free choice of level-0 models. They may either be derived from a single learning algorithm, or from a variety of different algorithms. The example in Section 3 uses different types of learning algorithms, while *bag-stacking*—stacking bagged models (Ting & Witten, 1997)—uses data variation to obtain a diverse set of models from a single learning algorithm. In the former case, performance may vary substantially between the level-0 models—for example NB performs very poorly in the Vowel and Euthyroid datasets compared to the other two models (see Table 2). Stacking copes well with this situation. The performance variation among the member models in bagging is rather small because they are derived from the same learning algorithm using bootstrap samples. Section 3.3 shows that a small performance variation among member models is a necessary condition for majority vote (as employed by bagging) to work well.

---

4. The heart dataset used by Breiman (1996b; 1996c) is omitted because it was very much modified from the original one.

It is worth noting that arcing and bagging can be incorporated into the framework of stacked generalization by using arced or bagged models as level-0 models. Ting & Witten (1997) show one possible way of incorporating bagged models with level-1 learning, employing MLR instead of voting. In this implementation, $\mathcal{L}$ is used as a *test* set for each of the bagged models to derive level-1 data rather than the cross-validated data. This is viable because each bootstrap sample leaves out about 37% of the examples. Ting & Witten (1997) show that bag-stacking almost always has higher predictive accuracy than bagging models derived from either C4.5 or NB. Note that the only difference here is whether an adaptive level-1 model or a simple majority vote is employed.

According to Breiman (1996b; 1996c), arcing and bagging can only improve the predictive accuracy of learning algorithms that are 'unstable.'[5] An unstable learning algorithm is one for which small perturbations in the training set can produce large changes in the derived model. Decision trees and neural networks are unstable; NB and IB1 are stable. Stacking works with both.

While MLR is the most successful candidate for level-1 learning that we have found, other algorithms might work equally well. One candidate is neural networks. However, we have experimented with back-propagation neural networks for this purpose and found that they have a much slower learning rate than MLR. For example, MLR only took 2.9 seconds as compare to 4790 seconds for the neural network in the nettalk dataset; while both have the same error rate. Other possible candidates are the multinomial logit model (Jordan & Jacobs, 1994), which is a special case of generalized linear models (McCullagh & Nelder, 1983), and the supra Bayesian procedure (Jacobs, 1995) which treats the level-0 models' confidence as data that may be combined with prior distribution of level-0 models via Bayes' rule.

## 5. Related Work

Our analysis of stacked generalization was motivated by that of Breiman (1996a), discussed earlier, and LeBlanc & Tibshirani (1993). LeBlanc & Tibshirani (1993) examine the stacking of a linear discriminant and a nearest neighbor classifier and show that, for one artificial dataset, a method similar to MLR performs better with non-negativity constraints than without. Our results in Section 3.2 show that these constraints are irrelevant to MLR's predictive accuracy in the classification situation.

LeBlanc & Tibshirani (1993) and Ting & Witten (1997) use a version of MLR that employs all class probabilities from each level-0 model to induce each linear regression. In this case, the linear regression for class $\ell$ is

$$LR_\ell(x) = \sum_k^K \sum_i^I \alpha_{ki\ell} P_{ki}(x).$$

This implementation requires the fitting of $KI$ parameters, as compared to $K$ parameters for the version used in this paper (see the corresponding formula in Section 2.2). Both versions give comparable results in terms of predictive accuracy, but the version used in this paper runs considerably faster because it needs to fit fewer parameters.

---

5. Schapire, R.E., Y. Freund, P. Bartlett, & W.S. Lee (1997) provide an alternative explanation for the effectiveness of arcing and bagging.

The limitations of MLR are well-known (Duda & Hart, 1973). For a $I$-class problem, it divides the description space into $I$ convex decision regions. Every region must be singly connected, and the decision boundaries are linear hyperplanes. This means that MLR is most suitable for problems with unimodal probability densities. Despite these limitations, MLR still performs better as a level-1 generalizer than IB1, its nearest competitor in deriving $\bar{\mathcal{M}}'$. These limitations may hold the key to a fuller understanding of the behavior of stacked generalization. Jacobs (1995) reviews linear combination methods like that used in MLR.

Previous work on stacked generalization, especially as applied to classification tasks, has been limited in several ways. Some only applies to a particular dataset (e.g., Zhang, Mesirov & Waltz, 1992). Others report results that are less than convincing (Merz, 1995). Still others have a different focus and evaluate the results on just a few datasets (LeBlanc & Tibshirani, 1993; Chan & Stolfo, 1995; Kim & Bartlett, 1995; Fan *et al.*, 1996).

One might consider a degenerate form of stacked generalization that does not use cross-validation to produce data for level-1 learning. Then, level-1 learning can be done 'on the fly' during the training process (Jacobs *et al.*, 1991). In another approach, level-1 learning takes place in batch mode, after all level-0 models are derived (Ho *et al.*, 1994).

Several researchers have worked on a still more degenerate form of stacked generalization without any cross-validation or learning at level 1. Examples are neural network ensembles (Hansen & Salamon, 1990; Perrone & Cooper, 1993; Krogh & Vedelsby, 1995), multiple decision tree combination (Kwok & Carter, 1990; Buntine, 1991; Oliver & Hand, 1995), and multiple rule combination (Kononenko & Kovačič, 1992). The methods used at level 1 are majority voting, weighted averaging and Bayesian combination. Other possible methods are distribution summation and likelihood combination. There are various forms of re-ordering class rank, and Ali & Pazzani (1996) study some of these methods for a rule learner. Ting (1996) uses the confidence of each prediction to combine a nearest neighbor classifier and a Naive Bayesian classifier.

## 6. Conclusions

We have addressed two crucial issues for the successful implementation of stacked generalization in classification tasks. First, class probabilities should be used instead of the single predicted class as input attributes for higher-level learning. The class probabilities serve as the confidence measure for the prediction made. Second, the multi-response least squares linear regression technique should be employed as the high-level generalizer. This technique provides a method of combining level-0 models' confidence. The other three learning algorithms have either algorithmic limitations or are not suitable for aggregating confidences.

When combining three different types of learning algorithms, this implementation of stacked generalization was found to achieve better predictive accuracy than both model selection based on cross-validation and majority vote; it was also found to be competitive with arcing and bagging. Unlike stacked regression, non-negativity constraints in the least-squares regression are not necessary to guarantee improved predictive accuracy in classification tasks. However, these constraints are still preferred because they increase the interpretability of the level-1 model.

The implication of our successful implementation of stacked generalization is that earlier model combination methods employing (weighted) majority vote, averaging, or other computations that do not make use of level-1 learning, can now apply this learning to improve their predictive accuracy.

## Acknowledgment

The authors are grateful to the New Zealand Marsden Fund for financial support for this research. This work was conducted when the first author was in Department of Computer Science, University of Waikato. The authors are grateful to J. Ross Quinlan for providing C4.5 and David W. Aha for providing IB1. The anonymous reviewers and the editor have provided many helpful comments.